\documentclass[runningheads]{llncs}

\usepackage[T1]{fontenc}
\usepackage{xcolor}%
\usepackage{times}
\usepackage{graphicx}
\usepackage{mdframed}
\usepackage{manyfoot}%

\begin{document}

\title{Roadmap on Incentive Compatibility for AI Alignment and Governance in Sociotechnical Systems}

%
\titlerunning{Roadmap on IC for AI Alignment and Governance in Sociotechnical Systems}

\author{Zhaowei Zhang\inst{1,2} \and
Fengshuo Bai\inst{3,4} \and
Mingzhi Wang\inst{1} \and
Haoyang Ye\inst{1} \and
Chengdong Ma\inst{1} \and
Yaodong Yang\inst{1}
}
\authorrunning{Z. Zhang, F. Bai et al.}
\institute{$^1$Institute for Artificial Intelligence, Peking University \\
$^2$State Key Laboratory of General Artificial Intelligence, BIGAI \\
$^3$Shanghai Jiao Tong University \quad $^4$Zhongguancun Academy \\
\email{zwzhang@stu.pku.edu.cn} \quad \email{yaodong.yang@pku.edu.cn}}

%
\maketitle              

\begin{abstract} 


The burgeoning integration of artificial intelligence (AI) into human society brings forth significant implications for societal governance and safety. While considerable strides have been made in addressing AI alignment challenges, existing methodologies primarily focus on technical facets, often neglecting the intricate sociotechnical nature of AI systems, which can lead to a misalignment between the development and deployment contexts.
To this end, we posit a new problem worth exploring: \textbf{I}ncentive \textbf{C}ompatibility \textbf{S}ociotechnical \textbf{A}lignment \textbf{P}roblem (ICSAP). We hope this can call for more researchers to explore how to leverage the principles of Incentive Compatibility (IC) from game theory to bridge the gap between technical and societal components to maintain AI consensus with human societies in different contexts.
We further discuss three classical game problems for achieving IC: mechanism design, contract theory, and Bayesian persuasion, in addressing the perspectives, potentials, and challenges of solving ICSAP, and provide preliminary implementation conceptions.

\end{abstract}

\section{Introduction}


The rapid development of artificial intelligence (AI) has had a significant impact on human society \cite{makridakis2017forthcoming,peeters2021hybrid,wamba2021we,tessler2024ai,zhang2025eurocon}, from robots entering human production and living environments \cite{michaelis2018reading,fu2024mobile} to large language models (LLMs) capable of complex natural language interactions \cite{zhao2023survey,bubeck2023sparks} and reasoning ability \cite{wei2022chain,wang2022self}. 
The problem will be much more significant for Artificial General Intelligence (AGI).
Therefore, an increasing number of people believe that as AI capabilities improve, AI systems will become integrated into human society in the future and be deployed in increasingly complex scenarios \cite{gladden2019will,dwivedi2021artificial}. Conversely, the powerful capabilities of AI systems have raised concerns about their safety \cite{cath2018artificial,peeters2021hybrid}, especially considering their behavioral motivations \footnote{https://www.scai.gov.sg/scai-question-6/}, alignment science \footnote{https://www.anthropic.com/news/core-views-on-ai-safety\label{anthropic-research-dir}} and how they align with human values and intentions \cite{ji2023ai}. This is recognized as the ``AI Alignment'' problem. 

Substantial progress has been made in addressing AI alignment issues, especially in the forward alignment process \cite{ji2023ai}, which enables AI systems to have alignment capabilities \textsuperscript{\ref{anthropic-research-dir}}. The methods for this process can mainly be divided into two categories. The first category involves learning from feedback \cite{christiano2017deep,bai2022training,ouyang2022training}, and there have been some significant research topics, including preference modeling \cite{wirth2017survey}, policy learning \cite{ibarz2018reward}, and scalable oversight \cite{christiano2018supervising,irving2018ai,bai2022training,burns2023weak}. The second category focuses on resolving distributional shift \cite{di2022goal,ngo2022alignment} in learning, with notable subproblems including algorithmic interventions \cite{vapnik1991principles,krueger2021out,lubana2023mechanistic}, adversarial training \cite{goodfellow2014explaining,poursaeed2021robustness}, and cooperative training \cite{dafoe2020open,dafoe2021cooperative,kang2022non,dong2023symmetry,qin2024multi}.

However, these methods only consider the given alignment objectives, focusing solely on technical components such as dataset, architecture, and training algorithms, etc. \cite{weidinger2023sociotechnical}, overlooking the fact that AI systems are sociotechnical systems \cite{selbst2019fairness}. 
Some studies have indicated that relying solely on technical means will result in a sociotechnical gap between the model's development context and its actual deployment context \cite{selbst2019fairness,lazar2023ai,zhang2025amulet}, which is also detrimental to further social governance. 
Such examples are not uncommon in daily life.
ChatGPT, trained on internet data and fine-tuned through RLHF \cite{ouyang2022training}, requires prompt engineering for adaptation to individual needs, highlighting unaddressed challenges in existing alignment techniques.
Additionally, for sociotechnical systems, existing research is more concerned with only societal components like governance and evaluation methods \cite{dean2021axes,weidinger2023sociotechnical}. 
Thus, currently, there is a lack of means to simultaneously consider both technical and societal components, enabling AI systems themselves to maintain consensus with human society.


Incentive Compatibility (IC) \cite{hurwicz1972informationally}, derived from game theory, suggests that participants only need to pursue their true interests to reach optimal outcomes \cite{roughgarden2010algorithmic}. This concept leverages self-interested behavior, aligning actions with the game designer's goals \cite{groves1987incentive}. With IC, each agent can maintain private goal information acquired during pretraining. Only by reconstructing different environments and rules,agents can optimize their own objectives to achieve outcomes that meet the needs of human society in different contexts. Therefore, we believe that exploring the IC property for AI alignment problems in sociotechnical systems is a highly worthwhile research endeavor.

In this paper, we separate a new subproblem from AI alignment problems in sociotechnical systems, called \textbf{I}ncentive \textbf{C}ompatibility \textbf{S}ociotechnical \textbf{A}lignment \textbf{P}roblem (ICSAP), and based on this, we propose our main position:
\begin{mdframed}[backgroundcolor=gray!10, roundcorner=10pt]
\textit{\large Achieving incentive compatibility can simultaneously consider both technical and societal components in the forward alignment phase, enabling AI systems to keep consensus with human societies in different contexts.}
\end{mdframed}

\section{Motivation and Opportunity: A Brief Example}
\label{sec:motivation}
In this section, we will use a very simple example to demonstrate how IC works in addressing AI alignment issues in sociotechnical systems through mechanism design, which will be illustrated specifically in Section \ref{sec:AMD}.

\begin{figure}[t]
  \centering  \includegraphics[width=0.8\textwidth]{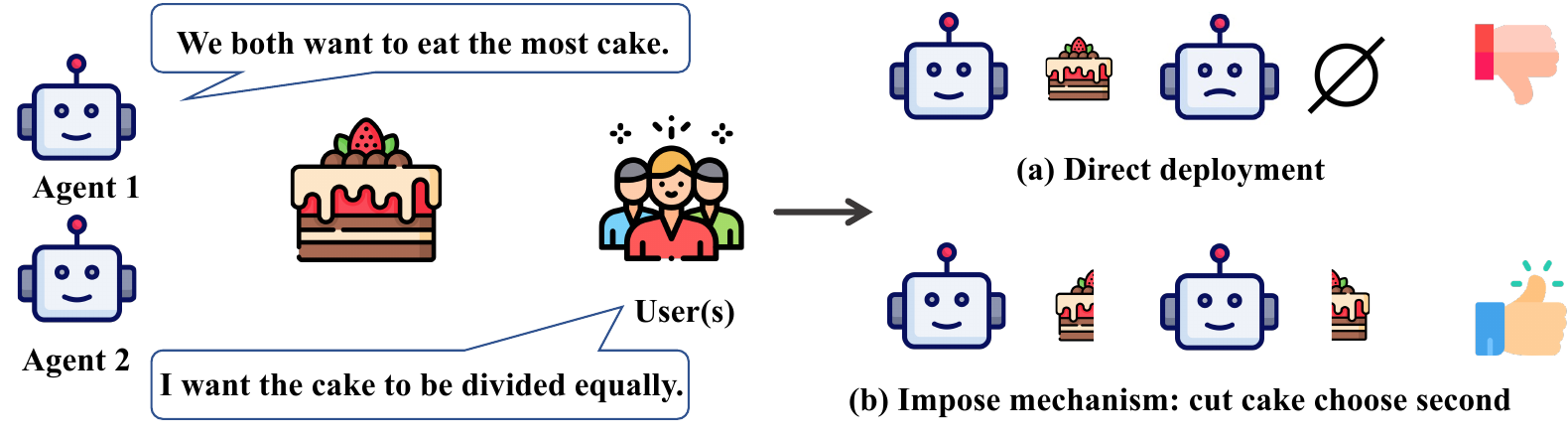} 
    \caption{
    A simple example illustrates how IC facilitates ICSAP scenarios through mechanism design. In the diagram, two agents aim to maximize cake consumption during technical training. However, the user desires equal cake distribution. Without IC, deploying both agents directly could lead to one party monopolizing the cake (a). With IC (b), the mechanism dictates that the second chooser is the one who cuts the cake. This ensures alignment with real-world needs by allowing agents to optimize within the rules, achieving the user's goal and aligning sociotechnical systems.
    }
  \label{fig:motivation_example}
\end{figure}


Consider a classic divide and choose problem: the two-player cake cutting \cite{steinhaus1948problem}. In this example (see Figure \ref{fig:motivation_example}), two self-interested agents aim to maximize their cake share, while the human seeks an equal division. If either agent cuts the cake, they'll take the whole. To align individual interests with the human's goal, a simple mechanism is proposed: the cutter chooses second. This constraint ensures the agent's pursuit of self-interest coincides with the host's objective, achieving Alignment. The mechanism's IC conditions facilitate consensus on an equal distribution and maximization of cake consumption.

If we consider the agent as an AI system and its desire to eat the most cake as the objective imparted by the technical component of training, we only need to use automated methods to search for corresponding mechanisms with IC properties as rules based on different real-world requirements to effectively solve ICSAP. 
Of course, hosts can also have different contextual needs, and they may not necessarily be self-interested. Here, we just provide a possible scenario to illustrate our point.

\section{Background and Overview}
In the following sections, we will demonstrate three classic game problems by applying media of IC properties: Mechanism Design in Section \ref{sec:AMD}, Contract Theory in Section \ref{sec:CT}, and Bayesian Persuasion in Section \ref{sec:BP}. An overview of these approaches is depicted in Figure \ref{fig:main_pic}.

\subsection{Mechanism Design}
\label{sec:AMD}

\begin{figure}[t]
  \centering  
\includegraphics[width=0.99\textwidth]{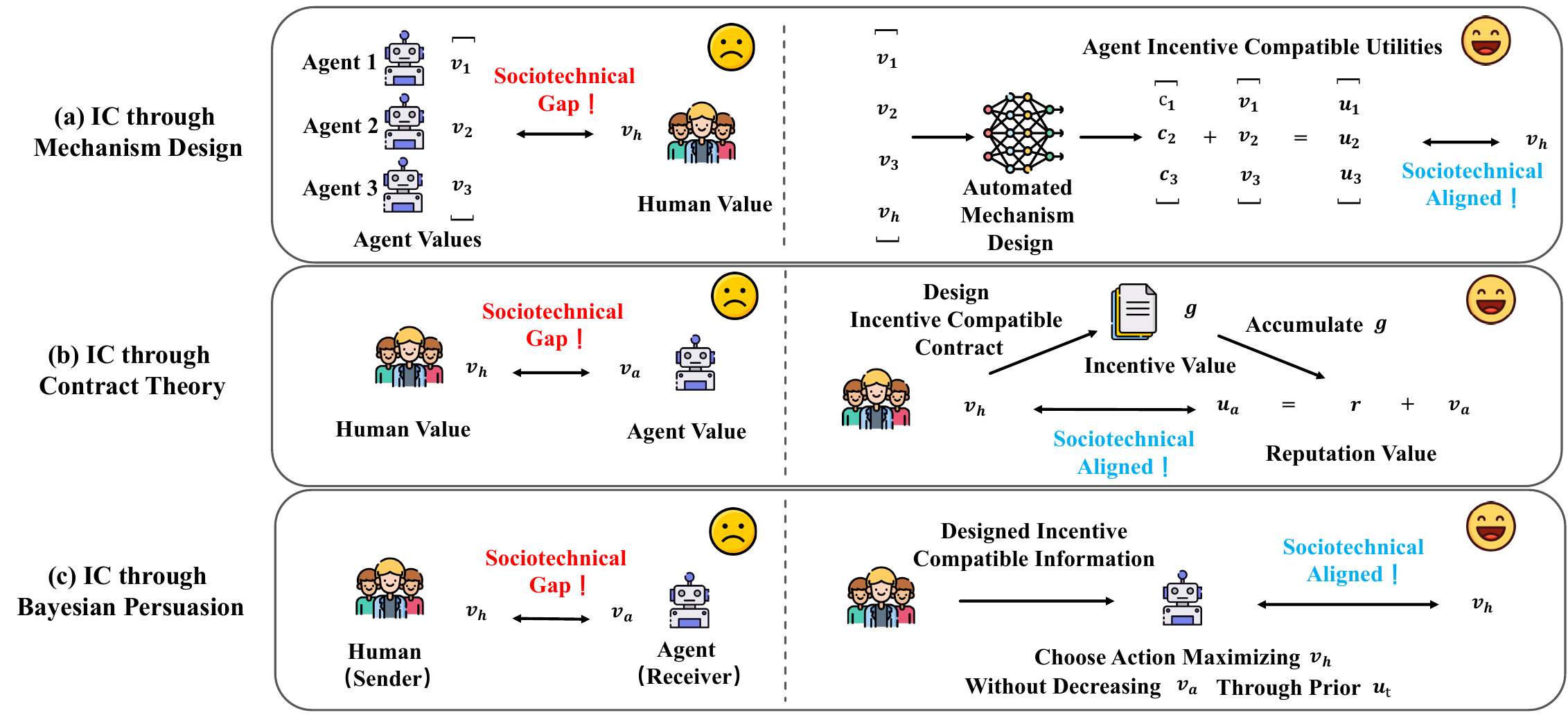}
    \caption{The figure illustrates how IC tackles ICSAP based on three classic game-theoretic problems.}
  \label{fig:main_pic}
\end{figure}

Mechanism Design theory deals with private information games where individual types and values are unknown to the designer \cite{nisan1999algorithmic}. 
It typically promotes heterogeneous value agents to reveal their private information and reach equilibrium at desired outcomes by constructing an efficient social structure for incentives \cite{dafoe2020open,ji2023ai}.

In mechanism design, IC is a fundamental constraint, alongside individual rationality, that restricts the possible mechanisms and social functions. However, the revelation principle \cite{Dasgupta1979TheIO} shows that IC doesn't limit our ability but simplifies strategic behaviors in rule design. It states that every Bayesian-Nash implementable social choice function can be achieved with incentive compatibility, treating IC as a ``free lunch'' scenario and allowing focus within this context.

Due to the generalized definition and objectives of mechanism design, it finds numerous applications in social choice theory \cite{gibbard1973,satterthwaite1975}, voting theory \cite{Dasgupta2020StrategyProofnessIO}, stable matching \cite{shapley1962college}, and auction theory \cite{Myerson1981OptimalAD,Clarke1971MultipartPO,wang2024gemnet}. 
For example, \cite{heidari2018fairness,huang2019veil,weidinger2023using} studied and analyzed the impact of the Veil of Ignorance mechanism \cite{rawls1971atheory} on social fairness and found that it promotes societal governance. \cite{sinha2015mechanism,zheng2020ai,zheng2022ai} ensure the maximization of social welfare and fairness through algorithmic learning of tax mechanisms.

Among them, the mechanism design has been most widely applied in the auction field.
For example, the second-price auction \cite{vickrey1961counterspeculation} is one of the simplest IC mechanisms. 
In a single-item environment, under the rule where the highest bidder pays the second-highest price, the weakly dominant strategy for bidders is to honestly reveal their valuation. 
In multi-item scenarios, achieving IC and maximizing social welfare generally rely on the Vickrey-Clarke-Groves (VCG) mechanism \cite{Clarke1971MultipartPO}. 
This mechanism aligns bidder utility maximization with social welfare maximization by initially paying each bidder the sum of the others' valuations, and then using a payment (utility) function based solely on the other bids to ensure IC. By setting the payment function to collect payments equal to the maximum social welfare when the bidder is absent, the designer ensures no net payment is needed, thus accounting for the externalities generated by the bidders.

Recent work \cite{mckee2023scaffolding,orzan2025cooperation} has constructed environments that encourage people to compete or cooperate through mechanism design. We can similarly apply this approach to AI governance in order to bridge the sociotechnical gap between humans and AI.

The subfigure (a) in Figure \ref{fig:main_pic} illustrates the case of IC through mechanism design. The left side of the figure demonstrates a sociotechnical gap between agents considering only technical components and the values of real humans. On the right side, it shows that by designing corresponding mechanisms according to different needs, we can adjust the values of agents, aligning their utility with human requirements under IC conditions, thus achieving alignment in sociotechnical systems.


\subsection{Contract Theory}
\label{sec:CT}

Contract theory \cite{bolton2004contract} is a field of economics that studies how various economic agents establish, manage, and reinforce their relationships and transactions through contracts. This theory focuses on the design and implementation of contracts, as well as their impact on individual behavior and overall social welfare. The core issues include the incompleteness of contracts \cite{pavlov2022optimal}), the problem of asymmetric information \cite{avraham2012private}, and how these issues lead to adverse selection and moral hazard \cite{guesnerie1989hidden}. Contract theory is significant for understanding and guiding practices in corporate governance, labor markets, insurance, financial markets, and legal applications.

In human-AI collaboration, contract theory is essential for aligning behaviors and values. It tackles information asymmetry \cite{lim2020hierarchical}, common in scenarios where human and AI capacities in information processing and decision-making differ. The method involves creating contractual terms that align AI's specific goals with human broader interests. This ensures AI actions benefit not just its own objectives but also the collective human interests, reducing risks like adverse selection and moral hazard from asymmetric information \cite{yan2018contract}. 
The key is designing mechanisms to align AI with human goals, ensuring mutual benefits despite differences in information and objectives. This strategic alignment resolves incentive issues and enhances coordination in human-AI interactions, leading to synergistic outcomes.

\cite{ivanov2024principal} has developed an agent capable of continuously interacting with contracts and the environment, thereby having greater potential to effectively coordinate and motivate humans and AI agents in real-world socio-economic environments.


\subsection{Bayesian Persuasion}\label{sec:BP}

IC emphasizes the importance of designing decision-making rules that encourage individuals to align their self-interested actions with broader goals. This concept plays a key role in Bayesian persuasion \cite{kamenica2011bayesian}, a strategy where senders, like policymakers \cite{alizamir2020warning} or marketers \cite{drakopoulos2021persuading}, selectively share information to shape the beliefs and choices of receivers, such as the public \cite{de2021informing} or consumer \cite{chen2020signalling}. This strategy is based on the Bayesian principle, where receivers update their beliefs based on the information provided. The sender's goal is to influence these beliefs by strategically transmitting information, guiding receivers towards decisions that meet the sender's aims. Thus, Bayesian persuasion is about more than just choosing what information to share; it's about aligning information transmission with the receivers' motivations to effectively influence their decisions toward the sender's goals.

Considering the solid theoretical foundation \cite{nguyen2021bayesian,bergemann2016information}, profound impact \cite{kamenica2019bayesian}, and extensive research across various fields \cite{castiglioni2020online,gan2022bayesian,hossain2024multi}, applying Bayesian persuasion to AI systems holds significant potential. Specifically, Bayesian persuasion can be utilized in interactions between humans (senders) and AI systems (receivers) within the context of artificial intelligence ethics and human-machine collaboration. In this setting, Bayesian persuasion can be seen as a tool to ensure that the behavior of AI systems aligns with the values and objectives of their human designers \cite{zhang2022forward}. This approach harnesses the principles of Bayesian persuasion to guide AI systems towards decisions and actions that reflect human ethics and goals, offering a promising avenue for integrating human values into AI decision-making processes.

Bayesian persuasion has great potential for AI governance. For example, with LLMs, we can use prompts for information design. In fact, recent work \cite{bai2024efficient} has attempted to use this method to achieve model-agnostic real-time alignment in the process of AI governance.

In the left half of subfigures (b) and (c) in Figure \ref{fig:main_pic}, both depict a sociotechnical gap between humans and a single agent. The right half of (b) demonstrates humans designing contracts that satisfy IC conditions based on specific needs, thereby adjusting the values of the agent through the contract. The right half of (c) illustrates a scenario of Bayesian persuasion where humans design information satisfying IC conditions according to their own needs, allowing agents to choose actions maximizing human demands without compromising their own values, thus solving ICSAP.

\section{Discussion: Potentials and Challenges}
\label{sec:discussion}

In this section, we delve into the IC through the integration of mechanism design, contract theory, and Bayesian persuasion into solving ICSAP, reflecting on the intertwined potentials and challenges as we endeavor to align AI systems with human values and objectives.

\subsection{Mechanism Design}
\paragraph{\textcolor{blue}{Potentials:}} \textbf{Mechanism design, particularly with its IC principle, emerges as a promising approach to steer AI behavior toward socially desirable outcomes.} Specifically, its reverse-engineering nature, which designs rules and incentives based on desired outcomes, is significantly enhanced by the advent of automated mechanism design fused with deep learning. This fusion offers a pathway to create context-specific mechanisms optimized for particular AI-human interaction scenarios. 


\paragraph{\textcolor{red}{Challenges:}} \textbf{Human values is complex in sociotechnical contexts.} The traditional assumptions of utility maximization and rationality, standard in mechanism design, may not fully apply to AI agents with behavioral patterns fundamentally distinct from human rationality. Moreover, the stability and robustness of mechanisms under variable conditions and their adaptability to complex social values like fairness and justice remain pressing concerns.

\subsection{Contract Theory}
\paragraph{\textcolor{blue}{Potentials:}} \textbf{Contract theory presents a unique framework for aligning AI with human values through self-enforcing contracts.} These contracts are tailored to intrinsically motivate AI towards actions that harmonize with human ethical standards. Incorporating incentive structures and reputation mechanisms, this theory addresses the critical issue of enforcing AI behavior, with potential implementation through neural networks to dynamically tune AI actions.


\paragraph{\textcolor{red}{Challenges:}} \textbf{Bridging the asymmetric information gap between AI and human intentions, and mitigating moral hazards where AI actions might deviate from ethical outcomes, are substantial.} These issues call for a strategic approach that combines a deep understanding of AI operations with the creation of robust and adaptable incentives to ensure AI behavior aligns consistently with human values.


\paragraph{\textcolor{red}{Challenges:}} \textbf{It is hard to overcome the gap between economic objectives and various real-world human requirements.} The challenge highlights the need for a more subtle approach to mechanism design in AI contexts, especially considering the limitations in the generalization capabilities of current automated design algorithms.

\subsection{Bayesian Persuasion}
\paragraph{\textcolor{blue}{Potentials:}} \textbf{Bayesian persuasion offers a nuanced avenue for influencing AI behavior by manipulating information structures.} This approach enables a dynamic interaction between human intentions and AI actions and will be particularly beneficial where direct control over AI is impractical, allowing for subtle yet effective steering of AI decisions. 


\paragraph{\textcolor{red}{Challenges:}} \textbf{Bayesian persuasion involves precise steps that make its effective implementation very difficult.} Challenges from this aspect are multifaceted, involving accurate modeling of belief systems, effective crafting of signal structures in partially observable environments, and bridging communication gaps between humans and AI. Addressing these challenges is crucial to effectively guide AI systems in a manner that aligns with human values, acknowledging the complexities and evolving nature of AI-human interactions.


\section{Conclusion}
\label{sec:conclusion}

In this paper, we highlight the sociotechnical gap between alignment research and real-world deployment, lacking effective means to address both technical and societal aspects simultaneously. We propose exploring IC for AI alignment and governance problems in sociotechnical systems as a valuable research pursuit. Our position argues that achieving IC can address both technical and societal components in the alignment phase, enabling AI systems to maintain consensus with human societies in various contexts. We use mechanism design, contract theory, and Bayesian persuasion to illustrate how our approach can bridge the sociotechnical gap.
Of course, this issue also faces many challenges, such as how to define complex human needs in sociotechnical scenarios. In future research, we call for more researchers to pay attention to this issue and propose more solutions from the perspective of ICSAP.

\section{Acknowledgments}

The work of ICSAP was supported by the National Natural Science Foundation of China (62376013).


\newpage
\bibliographystyle{splncs04}
\bibliography{main}


\end{document}